\documentclass[10pt,twocolumn,letterpaper]{article}

\usepackage{cvpr}
\usepackage{times}
\usepackage{epsfig}
\usepackage{graphicx}
\usepackage{amsmath}
\usepackage{amssymb}
\usepackage{colortbl}
\usepackage{vcell}
\usepackage{multirow}
\usepackage{arydshln}

\usepackage[pagebackref,breaklinks,colorlinks,bookmarks=false]{hyperref}

\begin{document}

\title{Exploring Self-Supervised Vision Transformers for Deepfake Detection: \\A Comparative Analysis}

\author{Huy H. Nguyen$^1$, Junichi Yamagishi$^{1,2}$, and Isao Echizen$^{1,2,3}$ \\
$^1$National Institute of Informatics, Japan\ \ \ \ \ \ $^2$SOKENDAI, Japan\ \ \ \ \ \ $^3$The University of Tokyo, Japan\\
{\tt\small \{nhhuy, jyamagis, iechizen\}@nii.ac.jp}
}

\maketitle
\thispagestyle{empty}

\begin{abstract}
This paper investigates the effectiveness of self-supervised pre-trained vision transformers (ViTs) compared to supervised pre-trained ViTs and conventional neural networks (ConvNets) for detecting facial deepfake images and videos. It examines their potential for improved generalization and explainability, especially with limited training data. Despite the success of transformer architectures in various tasks, the deepfake detection community is hesitant to use large ViTs as feature extractors due to their perceived need for extensive data and suboptimal generalization with small datasets. This contrasts with ConvNets, which are already established as robust feature extractors. Additionally, training ViTs from scratch requires significant resources, limiting their use to large companies. Recent advancements in self-supervised learning (SSL) for ViTs, like masked autoencoders and DINOs, show adaptability across diverse tasks and semantic segmentation capabilities. By leveraging SSL ViTs for deepfake detection with modest data and partial fine-tuning, we find comparable adaptability to deepfake detection and explainability via the attention mechanism. Moreover, partial fine-tuning of ViTs is a resource-efficient option.
\end{abstract}

\vspace{-1mm}

\section{Introduction}
In recent years, facial deepfake detection has emerged as a highly investigated field driven by the increasing prevalence of synthetic media~\cite{cardenuto2023age}. Transfer learning, a commonly adopted strategy in computer vision, has also been widely used in deepfake detection~\cite{tolosana2020deepfakes, rana2022deepfake}. The selection of an appropriate backbone architecture plays an important role, serving not only as a feature extractor but also as a regularizer to prevent overfitting. Previous studies have predominantly relied on ConvNets that are pre-trained using supervised learning on ImageNet. However, with the recent advancements in transformer architectures~\cite{vaswani2017attention}, such as CLIP~\cite{radford2021learning} and GPT-4~\cite{achiam2023gpt}, particularly in multi-modal tasks, there has been growing interest in exploring their efficacy for deepfake detection. Despite their demonstrated success in various domains, the adoption of pre-trained ViTs~\cite{dosovitskiy2020image} as feature extractors, especially large ones, has thus far been met with hesitation in the deepfake detection community. This reluctance stems from concerns about their immense capacity, which may exceed the requirements of the task and lead to potential overfitting, as well as their demanding resource requirements regarding training or fine-tuning data and computational resources. In contrast, ConvNets have already established themselves as robust feature extractors.

The advent of SSL has revolutionized the field of transformers, beginning with natural language processing (NLP) models such as BERT~\cite{kenton2019bert} and GPT~\cite{radford2019language}. Subsequently, DINO~\cite{caron2021emerging} and masked autoencoders (MAEs)~\cite{he2022masked} showcased the successful adaptation of SSL on ViTs, resulting in robust feature extractors and enabling explicit semantic segmentation of images---an ability not readily available in supervised ViTs. The introduction of registers~\cite{darcet2024vision} in DINOv2~\cite{oquab2023dinov2} further validated these capabilities, demonstrating their effectiveness in transfer learning across various downstream tasks. Specifically, in the realm of deepfake detection, the preliminary work of Cocchi \textit{et al.}~\cite{cocchi2023unveiling} demonstrated that detectors using either a basic k-NN classifier or a linear classifier equipped with pre-trained frozen DINO backbones can effectively identify images generated by Stable Diffusion models~\cite{rombach2022high}.

The current study substantially expands upon the findings of Cocchi \textit{et al.}~\cite{cocchi2023unveiling} in many key aspects. With the focused on comparing various backbones as feature extractors, \textit{i.e.}, evaluating the quality of their extracted representations for deepfake detection, our contributions can be summarized as follows:
\begin{itemize}
\item We conducted a comparative study on using pre-trained ViTs in facial deepfake detection from two perspectives: using their frozen backbones as multi-level feature extractors and partially fine-tuning their final transformer blocks. We employed simple classifiers as a proof-of-concept, which offers two benefits: reducing nuisance factors that could influence the comparison and ensuring that the results can generalize to any downstream classifier, whether simple or complex.

\item We highlight the advantages of partially fine-tuning the final blocks, demonstrating improvements in performance and natural explainability of the detection result via the attention mechanism, despite being fine-tuned on a small dataset with binary class annotations.
\item We conclude that leveraging SSL on ViTs, particularly DINOs, pre-trained using large datasets unrelated to deepfake detection, leads to a superior performance on the detection of various deepfakes compared to utilizing supervised pre-training.
\end{itemize}

\section{Related Work}
\subsection{Self-supervised vision transformers}
Caron \textit{et al.}~\cite{caron2021emerging} argued that image-level supervision oversimplifies rich visual information to a single concept. They applied SSL on a ViT~\cite{dosovitskiy2020image} called DeiT~\cite{touvron2021training} to improve feature representation, resulting in DINO. This approach can be seen as a type of knowledge distillation~\cite{hinton2015distilling} without explicit labels, leveraging techniques such as a momentum encoder~\cite{he2020momentum}, multi-crop training~\cite{caron2020unsupervised}, and small patches with ViTs. Subsequently, He \textit{et al.} used a masking mechanism for training MAEs, similar to BERT's~\cite{kenton2019bert}. SSL training with DINO and MAE showcased remarkable performance across various vision tasks and transfer learning via fine-tuning on downstream tasks. Importantly, DINO offered explicit information about semantic segmentation, a capability not clearly present in supervised ViTs.

The introduction of DINOv2~\cite{oquab2023dinov2} focused on accelerating and stabilizing training at scale using a larger dataset from curated and uncurated sources. Darcet \textit{et al.}~\cite{darcet2024vision} identified artifacts in the feature maps of supervised and self-supervised ViTs and proposed a solution that augments the input sequence with additional tokens, called registers. These registers are used during training but discarded during inference. These enhancements improve the robustness and efficiency of training self-supervised DINO, facilitating their application in various computer vision tasks, including deepfake detection.

\subsection{Transformers in deepfake detection}

In deepfake detection, transformers are primarily used in two ways: as feature refiners following ConvNets or as replacements for ConvNets as the main feature extractors.

The use of transformers as feature refiners has gained popularity since the introduction of the transformer architecture. In this approach, a ConvNet or an ensemble of ConvNets, often pre-trained and sometimes partially fine-tuned, serves as the primary feature extractor. A transformer, typically shallow with few blocks, is then trained to refine the extracted features. Khan \textit{et al.}~\cite{khan2021video} used XceptionNet~\cite{chollet2017xception} for feature extraction, followed by 12 transformer blocks for feature refining. Similarly, Wang \textit{et al.}~\cite{wang2022m2tr} used EfficientNet-B4~\cite{tan2019efficientnet} as a feature extractor and introduced a multi-scale transformer as one branch while utilizing a frequency filter as another branch for additional processing of the extracted features. Coccomini \textit{et al.}~\cite{coccomini2022combining} opted for the smaller extractor (EfficientNet-B0). Wang \textit{et al.}~\cite{wang2023deep} used ConvNets followed by a proposed convolutional pooling transformer before classification. Lin \textit{et al.}~\cite{lin2023deepfake} used a ConvNet for pre-processing, followed by a two-stream ConvNet and a transformer-based module. Notably, Zhao \textit{et al.}~\cite{zhao2023istvt} used XceptionNet as a feature extractor and proposed the interpretable spatial-temporal ViT with decomposed spatial-temporal self-attention and a self-subtract mechanism to capture spatial artifacts and temporal inconsistency.

The second way of using transformers as the main feature extractors involves proposing novel architectures or leveraging the pre-trained large ViTs. Heo \textit{et al.}~\cite{heo2023deepfake} combined patch embedding with EfficientNet-B7's features for pre-processing and applied a distillation method of DeiT~\cite{touvron2021training} on a transformer for deepfake detection. Guan \textit{et al.}~\cite{guan2022delving} introduced a local sequence transformer to model temporal consistency on sequences of restricted spatial regions. Ojha \textit{et al.}~\cite{ojha2023towards} implemented nearest neighbor and linear probing on the frozen supervised pre-trained CLIP's ViT~\cite{radford2021learning} for deepfake detection. Cocchi \textit{et al.}~\cite{cocchi2023unveiling} used the same classifiers as Ojha \textit{et al.} but additionally evaluated the self-supervised pre-trained DINO and DINOv2 as feature extractors. Liu \textit{et al.}~\cite{liu2024forgery} developed a forgery-aware adapter integrated into a frozen CLIP's ViT, adapting image features to discern and integrate local forgery traces within image and frequency domains. Das \textit{et al.}~\cite{das2024limited} borrowed the concept of MAE for the self-supervised auxiliary task. Motivated by the work of Ojha \textit{et al.}, Cocchi \textit{et al.}, and Liu \textit{et al.}~\cite{ojha2023towards, cocchi2023unveiling, liu2024forgery}, we conduct a comparative analysis on the use of self-supervised ViTs for deepfake detection from two perspectives: 1) utilizing their frozen backbones as multi-level feature extractors and 2) partially fine-tuning their final transformer blocks.

\begin{figure}[t]
\centering
\includegraphics[width=\columnwidth]{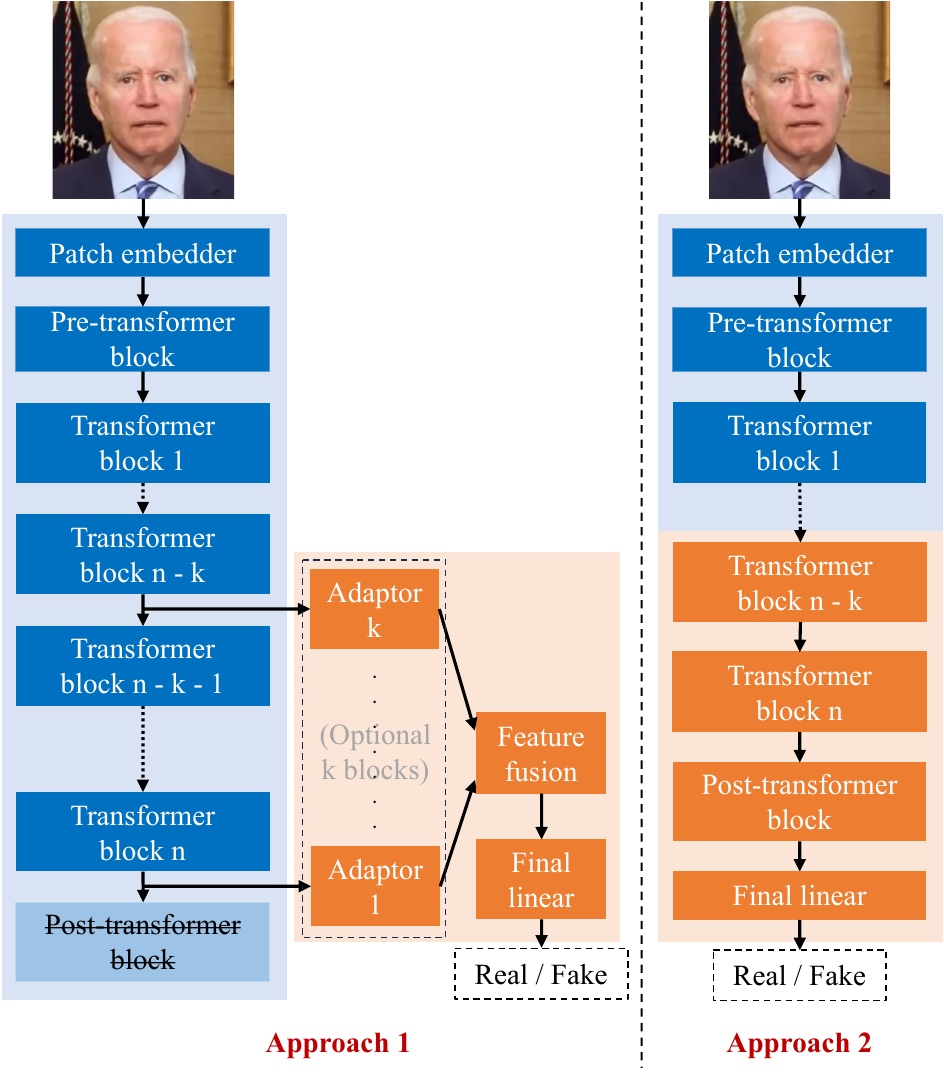}
\caption{Overview of the two investigated approaches. Blue blocks mean frozen blocks, while orange blocks mean fine-tuned or trained blocks.}
\label{fig:overview}
\end{figure}

\section{Methodology}

We introduce two approaches leveraging pre-trained ViTs, as illustrated in Fig.~\ref{fig:overview}. \textbf{Approach 1} is a generalized form that uses a frozen backbone as a feature extractor, widely adopted in deepfake detection~\cite{tolosana2020deepfakes, rana2022deepfake} and often paired with sophisticated adaptors~\cite{liu2024forgery}. This strategy is endorsed in the original DINO and MAE papers for various classification tasks~\cite{caron2021emerging, he2022masked} and in detecting generative adversarial network (GAN) and diffusion images~\cite{ojha2023towards, cocchi2023unveiling}. \textbf{Approach 2} involves fine-tuning the final transformer blocks, a less common technique due to the perception of transformers having large capacities and many parameters.

\subsection{Problem formalization}
As a basic binary classification problem, given an input image $I$ and a pre-trained backbone $\mathcal{B}$ with the classifier head removed, the objective is to construct a network $\mathcal{F}$ utilizing $\mathcal{B}$ to classify $I$ as either ``real'' or ``fake'', corresponding to the binary labels $0$ or $1$. This can be represented as
\begin{equation}
\text{output} = \begin{cases} 
1 & \text{if } \sigma(\mathcal{F}(\mathcal{B}(I))) \geq \tau \\
0 & \text{otherwise},
\end{cases}
\end{equation}
where $\sigma(\cdot)$ is the sigmoid function, mapping the output of $\mathcal{F}(\mathcal{B}(I))$ to a probability in the range $[0, 1]$, and $\tau$ is the threshold value. Note that the softmax function could be used instead of sigmoid to convert the logits extracted by $\mathcal{F}$ into probabilities, but utilizing softmax makes it easier to extend from binary to multi-class classification.

The backbone $\mathcal{B}$ begins with a pre-processing module and consists of $n$ blocks. For simplicity, we denote the intermediate features of $I$ extracted by block $i$ as $\phi_i$.

Regarding the value of $\tau$, there are various methods for determining its optimal value, which can vary from one paper to another. In this work, we set $\tau$ to either 0.5 or to the threshold corresponding to the equal error rate (EER) calculated on the validation set, depending on the experiment settings.

\begin{table*}[t]
\centering
\caption{Backbones used in the experiments.}
\label{tab:backbones}
\resizebox{0.98\textwidth}{!}{
\begin{tabular}{l|l|l|lrl}
\hline
\textbf{Backbone} & \textbf{Architecture} & \textbf{Way of training} & \textbf{Dataset(s)} & \textbf{Images} & \textbf{Annotations} \\ 
\hline
EfficientNetV2 Large~\cite{tan2021efficientnetv2} & ConvNet & Supervised & ImageNet-21K~\cite{deng2009imagenet} & 14M & Image classes \\ 
\hdashline
DeiT III L/16-LayerScale~\cite{touvron2022deit} & Transformer & Supervised & ImageNet-21K~\cite{deng2009imagenet} & 14M & Image classes \\ 
EVA-02-CLIP-L/14~\cite{sun2023eva} & Transformer & Supervised & \begin{tabular}[c]{@{}l@{}}LAION-2B~\cite{schuhmann2022laion} \& \\COYO-700M~\cite{kakaobrain2022coyo-700m}\end{tabular} & 2B & Image-text pairs \\ 
\hdashline
MAE ViT-L/16~\cite{he2022masked} & Transformer & Self-supervised & ImageNet-1K~\cite{deng2009imagenet} & 1.3M & Not used \\
DINO (various versions)~\cite{caron2021emerging} & Transformer & Self-supervised & ImageNet-21K~\cite{deng2009imagenet} & 14M & Not used \\
DINOv2 (various versions)~\cite{oquab2023dinov2, darcet2024vision} & Transformer & Self-supervised & LVD-142M~\cite{oquab2023dinov2} & 142M & Not used \\
\hline
\end{tabular}}
\end{table*}

\subsection{Approach 1: Using frozen backbone as a multi-level feature extractor}
In this approach, intermediate features $\phi_i$ can be further processed by an adaptor $\mathcal{A}$ (optional) before being fused with other intermediate features extracted by other blocks through a feature fusion operation $\Sigma$, followed by a classifier $\mathcal{C}$, typically a linear one. This approach is visualized on the left part of Fig.~\ref{fig:overview}. The backbone $\mathcal{B}$ remains frozen. We utilize the $k$ final intermediate features extracted by the $k$ final blocks. This can be formalized by 

\begin{equation}
\mathcal{F}(\mathcal{B}(I)) = \mathcal{C}\left(\sum_{i=n-k}^{n} \mathcal{A}_i(\phi_i)\right).
\end{equation}

Due to the nature of the comparative study, we opted not to use complex architectures for the adaptor $\mathcal{A}$ and the classifier $\mathcal{C}$ here. Instead, we utilized dropout and linear layers for their construction. Further details can be found in Sec.~\ref{sec:res_n_dis}.

\subsection{Approach 2: Fine-tuning last transformer blocks}
This approach is more straightforward than \textbf{Approach 1}. We append the new classifier $\mathcal{C}$ after the backbone $\mathcal{B}$, as visualized on Fig.~\ref{fig:overview}'s right part. This can be formalized by 
\begin{equation}
\mathcal{F}(\mathcal{B}(I)) = \mathcal{C}(\mathcal{B}(I)).
\end{equation}

During fine-tuning, the first $n-k$ blocks are frozen. The class (CLS) token and register tokens (if exist) are also fine-tuned along with the last $k$ blocks and the new classifier $\mathcal{C}$.

There are two major advantages of this approach compared to \textbf{Approach 1}:
\begin{itemize}
\item There are no additional parameters for the adaptors $\mathcal{A}$s and the feature fusion operation $\Sigma$. Given the already substantial size of recent feature extractors, particularly transformers, avoiding additional parameters is advantageous.
\item \textit{(For transformer backbones only)} Since the final transformer block and the tokens are fine-tuned, the attention weights to the CLS token are adapted to deepfake detection. They can be used naturally to visualize the focused area, similar to the visualization techniques used in the DINO papers~\cite{caron2021emerging, oquab2023dinov2, darcet2024vision}. This enhancement improves the detector's explainability, a crucial factor in deepfake detection.
\end{itemize}

\section{Experimental Design}
\subsection{Backbones}
We meticulously selected ConvNet and transformer backbones widely utilized in the domains of computer vision and forensics, prioritizing models with comparable sizes to ensure equitable and informative comparisons. Detailed specifications are provided in Tab.~\ref{tab:backbones}.

For ConvNets' representative, we opted for EfficientNetV2~\cite{tan2021efficientnetv2} due to its robust architecture and the popularity of its predecessor in the forensics community. For supervised ViTs, we selected two well-known models: DeiT III~\cite{touvron2022deit} (having the same architecture as DINO, using the conventional class labels annotation) and EVA-CLIP~\cite{sun2023eva} (an enhanced version of the renowned CLIP~\cite{radford2021learning}, using multimodal image-text pairs annotation). For SSL ViT, we chose DINOs~\cite{caron2021emerging, oquab2023dinov2, darcet2024vision} and MAE~\cite{he2022masked}, focusing more on DINOs for simplicity. We utilized the official pre-trained weights provided by the authors.

\subsection{Datasets}
We followed the data design of Nguyen \textit{et al.}~\cite{nguyen2023close} by gathering a variety of images generated or manipulated by various deepfake methods to construct the main datasets. The details of the training, validation, and test sets are shown in Tab.~\ref{tab:datasets}. The datasets were designed to be balanced regarding the ratio of real and fake images and the number of images per training method, and were guaranteed not to overlap.

\begin{table}[t]
\centering
\caption{Sizes of the main training, validation (val), and test (seen) sets, inspired by Nguyen \textit{et al.}~\cite{nguyen2023close}, and of the unseen validation and test sets from Ț{\^a}nțaru \textit{et al.}~\cite{tantaru2024weakly}.}

\label{tab:datasets}
\resizebox{0.88\columnwidth}{!}{
\begin{tabular}{l|r r r}
\hline
\textbf{Type} & \textbf{Real} & \textbf{Fake} & \textbf{Total} \\ \hline
Training & 44,037 & 55,963 & 100,000 \\
Validation & 13,200 & 13,000 & 26,200\\
Test & 10,000 & 11,000 & 21,000 \\
\hdashline
Validation (unseen) & 1,900 & 10,700 & 12,600 \\
Test (unseen) & 900 & 3,600 & 4,500 \\
\hline
\end{tabular}
}
\end{table}

\textbf{Real} images were gathered from the VidTIMIT~\cite{sanderson2009multi}, VoxCeleb2~\cite{chung2018voxceleb2}, FaceForensics++ (FF++)~\cite{rossler2019faceforensics++}, Google DFD~\cite{googledfd}, Deepfake Detection Challenge Dataset (DFDC)~\cite{dolhansky2020deepfake}, and Celeb-DF~\cite{li2020celeb} datasets. One part of the \textbf{fake} images comprised images gathered from the FF++, Google DFD, Celeb-DF, DFDC, DeepfakeTIMIT (DF-TIMIT)~\cite{korshunov2018deepfakes}, and YouTube-DF (YT-DF)~\cite{kukanov2020cost} datasets. The other part were images generated by various GANs, including StarGAN~\cite{choi2018stargan}, StarGAN-v2~\cite{choi2020stargan}, RelGAN~\cite{wu2019relgan}, ProGAN~\cite{karras2018progressive}, StyleGAN~\cite{karras2019style}, and StyleGAN2~\cite{karras2020analyzing}.

For \textbf{cross-dataset evaluation}, we used the dataset constructed by Ț{\^a}nțaru \textit{et al.}~\cite{tantaru2024weakly}, which contains images generated or manipulated by diffusion-based methods. It is important to note that our training and validation sets (main dataset) above do not contain any diffusion images. The list of diffusion-based methods used here includes Perception Prioritized (P2)~\cite{choi2022perception}, Repaint-P2~\cite{choi2022perception, tantaru2024weakly}, Repaint-Latent Diffusion Model (LDM)~\cite{rombach2022high, tantaru2024weakly}, Large Mask Inpainting (LaMa)~\cite{suvorov2022resolution}, and Pluralistic~\cite{zheng2019pluralistic}.

Regarding the roles of the subsets, we used the training set for training or fine-tuning models and the validation sets for hyper-parameter selection, including the selection of the best checkpoints and determination of the EER thresholds, which were then used for testing. The test sets were used for evaluation and comparison.

\begin{table}[t]
\centering
\caption{Performances of four conventional classifiers on various DINO and DINOv2 architectures.}
\label{tab:classical_cmp}
\resizebox{\columnwidth}{!}{
\begin{tabular}{lc|cccc} 
\hline
\multicolumn{1}{c}{\begin{tabular}[c]{@{}c@{}}\textbf{ViT}\\\textbf{backbone}\end{tabular}} & \begin{tabular}[c]{@{}c@{}}\textbf{DINO}\\\textbf{version}\end{tabular} & \begin{tabular}[c]{@{}c@{}}\textbf{PCA + }\\\textbf{k-means}\end{tabular} & \textbf{k-NN} & \textbf{Linear} & \begin{tabular}[c]{@{}c@{}}\textbf{MLP}\\\textbf{(2 layers)}\end{tabular} \\ 
\hline
\textbf{S}/8 & 1 & 58.69 & 69.88 & 65.69 & 76.98 \\
\textbf{S}/16 & 1 & 58.90 & 70.01 & 73.10 & \textbf{80.10} \\
\textbf{S}/14 & 2 & \textbf{60.15} & \textbf{71.07} & \textbf{77.06} & 77.21 \\
\textbf{S}/14-Reg & 2 & 59.63 & 69.88 & 74.16 & 77.99 \\ 
\hline
\textbf{B}/8 & 1 & \textbf{59.25} & 70.79 & \textbf{78.40} & \textbf{81.83} \\
\textbf{B}/16 & 1 & 58.53 & \textbf{70.96} & 67.23 & 81.07 \\
\textbf{B}/14 & 2 & 55.25 & 69.67 & 75.84 & 77.62 \\
\textbf{B}/14-Reg & 2 & 54.90 & 69.36 & 77.67 & 76.29 \\
\hline
\end{tabular}}
\end{table}

\subsection{Metrics}
We used some or all of the following five to measure the performance of the detectors: classification accuracy, true positive rate (TPR), true negative rate (TNR), equal error rate (EER), and half total error rate (HTER).

\begin{table*}[t]
\centering
\caption{EERs of models with \textbf{Approach 1} utilizing various versions and architectures of DINO as backbones on the main (seen) test set. ``Failed'' indicates that the models failed to converge during training.}
\label{tab:approach_1}
\resizebox{0.95\textwidth}{!}{
\begin{tabular}{l|c|ccccc|ccc} 
\hline
\multicolumn{1}{c|}{\vcell{\textbf{Model}}} & \vcell{\begin{tabular}[b]{@{}c@{}}\textcolor[rgb]{0.122,0.137,0.157}{\textbf{Back-}}\\\textcolor[rgb]{0.122,0.137,0.157}{\textbf{bone}}\\\textcolor[rgb]{0.122,0.137,0.157}{\textbf{params}}\end{tabular}} & \vcell{\begin{tabular}[b]{@{}c@{}}\textbf{CLS token}\\\textbf{Final block}\end{tabular}} & \vcell{\begin{tabular}[b]{@{}c@{}}\textbf{CLS token}\\\textbf{4 blocks}\\\textbf{WS}\end{tabular}} & \vcell{\begin{tabular}[b]{@{}c@{}}\textbf{CLS token}\\\textbf{12 blocks}\\\textbf{WS}\end{tabular}} & \vcell{\begin{tabular}[b]{@{}c@{}}\textbf{CLS token}\\\textbf{4 blocks}\\\textbf{concat}\end{tabular}} & \vcell{\begin{tabular}[b]{@{}c@{}}\textbf{CLS token}\\\textbf{12 final blocks}\\\textbf{concat}\end{tabular}} & \vcell{\begin{tabular}[b]{@{}c@{}}\textbf{All tokens}\\\textbf{Final block}\end{tabular}} & \vcell{\begin{tabular}[b]{@{}c@{}}\textbf{All tokens}\\\textbf{4 blocks}\\\textbf{WS}\end{tabular}} & \vcell{\begin{tabular}[b]{@{}c@{}}\textbf{All tokens}\\\textbf{8 blocks}\\\textbf{WS}\end{tabular}} \\[-\rowheight]
\multicolumn{1}{c|}{\printcelltop} & \printcelltop & \printcelltop & \printcelltop & \printcelltop & \printcelltop & \printcelltop & \printcelltop & \printcelltop & \printcelltop \\ 
\hline
\rowcolor[rgb]{0.753,0.753,0.753} \textbf{DINO} & & \multicolumn{1}{l}{} & \multicolumn{1}{l}{} & \multicolumn{1}{l}{} & \multicolumn{1}{l}{} & \multicolumn{1}{l|}{} & \multicolumn{1}{l}{} & \multicolumn{1}{l}{} & \multicolumn{1}{l}{} \\
ViT-\textbf{S}/8 & 21M & 26.43 & 21.16 & 21.23 & 20.41 & 18.72 & 16.03 & Failed & Failed \\
ViT-\textbf{S}/16 & 21M & 22.73 & 20.22 & 19.91 & 19.97 & 18.21 & 13.99 & 14.35 & 14.22 \\
ViT-\textbf{B}/8 & 85M & \textbf{19.85} & \textbf{18.67} & \textbf{18.03} & \textbf{18.17} & \textbf{16.40} & 14.09 & 13.87 & 17.74 \\
ViT-\textbf{B}/16 & 85M & 20.62 & 19.52 & 19.24 & 18.43 & 17.35 & \textbf{13.95} & \textbf{13.52} & \textbf{13.95} \\ 
\hline
\rowcolor[rgb]{0.753,0.753,0.753} \textbf{DINOv2} & & \multicolumn{1}{l}{} & \multicolumn{1}{l}{} & \multicolumn{1}{l}{} & \multicolumn{1}{l}{} & \multicolumn{1}{l|}{} & \multicolumn{1}{l}{} & \multicolumn{1}{l}{} & \multicolumn{1}{l}{} \\
ViT-\textbf{S}/14 & 21M & 23.25 & 22.99 & 21.81 & 20.28 & 18.76 & 14.63 & 14.53 & Failed \\
ViT-\textbf{S}/14-Reg & 21M & 26.78 & 23.16 & 23.05 & 20.61 & 18.92 & 15.02 & 15.26 & Failed \\
ViT-\textbf{B}/14 & 86M & 23.08 & 18.41 & 18.44 & 17.02 & 16.39 & 13.44 & 13.29 & Failed \\
ViT-\textbf{B}/14-Reg & 86M & 23.69 & 20.27 & 20.08 & 19.63 & 19.32 & 14.37 & 13.84 & Failed \\
ViT-\textbf{L}/14 & 300M & 21.72 & \textbf{16.84} & 16.39 & \textbf{15.82} & 14.51 & 13.01 & 13.09 & Failed \\
ViT-\textbf{L}/14-Reg & 300M & 22.43 & 18.97 & 16.88 & 18.75 & 15.19 & 14.00 & 12.67 & 12.64 \\
ViT-\textbf{G}/14 & 1,100M & 20.82 & 19.48 & 16.17 & 18.77 & 14.72 & \textbf{11.67} & 12.48 & \textbf{11.72} \\
ViT-\textbf{G}/14-Reg & 1,100M & \textbf{19.81} & 18.02 & \textbf{15.68} & 17.76 & \textbf{14.18} & 12.40 & \textbf{12.12} & 12.46 \\
\hline
\end{tabular}}
\end{table*}

\section{Results and discussion}
\label{sec:res_n_dis}
In this section, we first discuss \textbf{Approaches 1} and \textbf{2} in Sections~\ref{sec:eval_approach_1} and~\ref{sec:eval_approach_2}, respectively. We compare the performances among different architectures and versions of DINO, as well as between DINOs versus ConvNets and other ViTs. Additionally, we implement improvements and conduct ablation studies on the best-performing architectures to gain further enhancements and insights. Next, we evaluate selected models on the unseen test set in Sec.~\ref{sec:cross_dataset}. Lastly, we visualize the attention maps of the fine-tuned DINOv2 model versus their original SSL pre-trained versions on real and deepfake image examples in Sec.~\ref{sec:visualization}.

\subsection{Approach 1: Using frozen backbone as a multi-level feature extractor}
\label{sec:eval_approach_1}
We initially validated the findings of Cocchi \textit{et al.}~\cite{cocchi2023unveiling} with some minor extensions by applying conventional classifiers to the features extracted by frozen SSL pre-trained DINO backbones (with small size (S) and base size (B)), specifically extracted from the CLS token. In addition to the nearest neighbors (k-NN) classifier and linear probing, we utilized k-means on the principal component analysis (PCA) features~\cite{nguyen2023close} and a two-layer perceptron classifier. The results are shown in Tab.~\ref{tab:classical_cmp}.

Regarding the PCA + k-means classifier, most models achieved accuracies of around 58\%, which is slightly better than random guessing. As k-means is an unsupervised clustering method, this outcome suggests a partial ability of these pre-trained backbones to separate deepfakes without additional modules. For k-NN and linear probing, due to the complexity of the test set, the average accuracy was approximately 70\%, which is lower than the results reported by Cocchi \textit{et al.} on their diffusion dataset. The addition of an extra linear layer (the MLP) resulted in performance improvements ranging from about 2\% to 14\% in most cases, suggesting the potential for further enhancements.

Deepfake detection involves identifying deepfake fingerprints, such as artifacts or irregular patterns. Therefore, relying solely on the CLS token may not be optimal. We assessed the effectiveness of incorporating patch tokens and multiple intermediate features from the final $k$ blocks rather than solely from the final block. We also compared two feature fusion techniques: weighted sum (WS) and concatenation (concat). Given the large feature sizes when using all tokens (CLS and patch tokens), we only evaluated the weighted sum technique in this case. For the DINO backbones, we evaluated small (S), base (B), large (L), and giant (G) sizes, the latter two being available only in DINOv2. The results are presented in Tab.~\ref{tab:approach_1}.

The principle of ``larger is better'' applies here, where larger backbone sizes generally result in lower EERs. Utilizing all tokens yields significantly better results than relying solely on the CLS token. Moreover, utilizing multiple blocks performs better than using a single block, and the performance further improves with larger values of $k$. However, training downstream modules becomes more challenging as $k$ increases, leading to convergence issues in some cases (denoted as ``Failed''). Concatenating features yields better results than utilizing a weighted sum. Across similar sizes, there is generally no discernible difference in performance between DINO and DINOv2. Notably, for DINO, there is no clear performance distinction between using large and small patch sizes.

\begin{table}[t]
\centering
\caption{Enhancements to \textbf{Approach 1} utilizing SSL pre-trained DINOv2 - ViT-L/14-Reg as the backbone, with ``L'' denoting linear adaptors. Accuracies were calculated using a threshold of 0.5.}
\label{tab:approach_1_abl}
\resizebox{\columnwidth}{!}{
\begin{tabular}{cccccc}
\hline
\textbf{Blocks} & \textbf{Dropout} & \textbf{Fusion} & \textbf{Accuracy} & \textbf{EER} & \textbf{HTER} \\ 
\hline
1 & No & -- & 85.48 & 14.00 & 14.28 \\
1 & \textbf{Yes} & -- & 86.26 & 13.57 & 13.97 \\ 
\hdashline
4 & No & WS & 84.15 & 12.67 & 15.35 \\
4 & \textbf{Yes} & WS & 85.31 & 12.38 & 14.27 \\
\hdashline
4 & No & L+concat & 86.52 & 13.41 & 13.35 \\
4 & \textbf{Yes} & L+concat & \textbf{87.42} & \textbf{11.98} & \textbf{12.41} \\
\hline
\end{tabular}}
\end{table}

\begin{table*}[t]
\centering
\caption{Comparison between different ConvNet and transformer architectures on the seen test set. For \textbf{Approach 1}, the final setting in Tab.~\ref{tab:approach_1_abl} was applied. $k$ denotes the number of final blocks utilized for feature extraction in \textbf{Approach 1} while signifies the number of fine-tuned blocks in \textbf{Approach 2}. Accuracies were calculated using a threshold of 0.5.}
\label{tab:comparison}
\resizebox{\textwidth}{!}{
\begin{tabular}{l|c:cccccccccccccccccc} 
\hline
\multicolumn{1}{c|}{\vcell{\textbf{Model}}} & \vcell{$\mathbf{k}$} & \vcell{\begin{tabular}[b]{@{}c@{}}\textbf{FF++}\\\textbf{Real}\end{tabular}} & \vcell{\begin{tabular}[b]{@{}c@{}}\textbf{FF++}\\\textbf{DF}\end{tabular}} & \vcell{\begin{tabular}[b]{@{}c@{}}\textbf{FF++}\\\textbf{F2F}\end{tabular}} & \vcell{\begin{tabular}[b]{@{}c@{}}\textbf{FF++}\\\textbf{FS}\end{tabular}} & \vcell{\begin{tabular}[b]{@{}c@{}}\textbf{FF++}\\\textbf{NT}\end{tabular}} & \vcell{\begin{tabular}[b]{@{}c@{}}\textbf{FF++}\\\textbf{FSh}\end{tabular}} & \vcell{\begin{tabular}[b]{@{}c@{}}\textbf{DFD}\\\textbf{Real}\end{tabular}} & \vcell{\begin{tabular}[b]{@{}c@{}}\textbf{DFD}\\\textbf{Fake}\end{tabular}} & \vcell{\begin{tabular}[b]{@{}c@{}}\textbf{Vid-}\\\textbf{TIMIT}\end{tabular}} & \vcell{\begin{tabular}[b]{@{}c@{}}\textbf{DF-}\\\textbf{TIMIT}\end{tabular}} & \vcell{\begin{tabular}[b]{@{}c@{}}\textbf{Vox-}\\\textbf{Celeb2}\end{tabular}} & \vcell{\begin{tabular}[b]{@{}c@{}}\textbf{YT-}\\\textbf{DF}\end{tabular}} & \vcell{\begin{tabular}[b]{@{}c@{}}\textbf{DFDC}\\\textbf{Real}\end{tabular}} & \vcell{\begin{tabular}[b]{@{}c@{}}\textbf{DFDC}\\\textbf{Fake}\end{tabular}} & \vcell{\textbf{GANs}} & \vcell{\textbf{\textbf{Acc.}}} & \vcell{\textbf{EER}} & \vcell{\textbf{HTER}} \\[-\rowheight]
\multicolumn{1}{c|}{\printcelltop} & \printcelltop & \printcellmiddle & \printcelltop & \printcelltop & \printcelltop & \printcelltop & \printcelltop & \printcelltop & \printcelltop & \printcelltop & \printcelltop & \printcelltop & \printcelltop & \printcelltop & \printcelltop & \printcelltop & \printcelltop & \printcelltop & \printcelltop \\ 
\hline
\rowcolor[rgb]{0.753,0.753,0.753} \multicolumn{1}{|l|}{\textbf{Approach 1}} &  &  &  &  &  &  &  &  &  &  &  &  &  &  &  &  &  &  &  \\
EfficientNetV2 Large & 4 & 60.90 & 84.70 & 83.40 & 79.50 & 75.30 & 83.50 & 61.00 & 85.50 & 59.75 & 67.90 & 92.62 & 67.40 & 54.00 & 71.60 & 95.55 & 78.50 & 21.71 & 21.62 \\
DeiT III L/16-LayerScale & 4 & 38.80 & 88.00 & 87.10 & 81.60 & 82.20 & 82.70 & 37.60 & 92.00 & 93.45 & 39.90 & 99.68 & 53.10 & 57.20 & 62.20 & 95.70 & 79.96 & 19.77 & 19.96 \\
EVA-02-CLIP-L/14~ & 4 & 50.40 & 97.40 & 90.80 & 93.20 & 80.80 & 90.30 & 44.50 & 97.20 & 83.30 & 66.40 & 99.82 & 31.90 & 57.10 & 83.90 & 99.90 & 83.30 & 16.51 & 16.77 \\ 
\hdashline
MAE ViT-L/16 & 4 & 47.20 & 97.50 & 93.70 & 94.20 & 91.00 & \multicolumn{1}{c|}{97.70} & 54.20 & 98.40 & 91.75 & 71.90 & 99.70 & 85.00 & 53.70 & 82.90 & 99.90 & 88.06 & \textbf{11.90} & \textbf{12.14} \\
DINOv2 ViT-L/14-Reg & 4 & 72.40 & 94.60 & 87.00 & 89.30 & 76.50 & 89.40 & 69.40 & 98.30 & 96.80 & 46.20 & 99.70 & 67.30 & 77.70 & 76.00 & 99.85 & 87.42 & \underline{11.98} & \underline{12.41} \\ 
\hline
\rowcolor[rgb]{0.753,0.753,0.753} \textbf{Approach 2} &  &  &  &  &  &  &  &  &  &  &  &  &  &  &  &  &  &  &  \\
EfficientNetV2 Large & 1 & 65.00 & 94.60 & 89.00 & 89.30 & 84.10 & 90.50 & 22.90 & 91.50 & 99.20 & 60.40 & 99.16 & 62.50 & 72.40 & 68.90 & 97.25 & 84.75 & 15.05 & 15.22 \\
DeiT III L/16-LayerScale & 1 & 56.90 & 97.00 & 89.50 & 89.90 & 84.10 & 86.50 & 33.40 & 95.00 & 96.35 & 18.60 & 99.08 & 56.50 & 66.40 & 79.00 & 97.25 & 82.64 & 17.21 & 17.28 \\
EVA-02-CLIP-L/14~ & 1 & 53.50 & 97.50 & 92.70 & 92.40 & 83.30 & 92.40 & 62.40 & 98.80 & 96.55 & 62.90 & 99.26 & 82.20 & 64.00 & 84.60 & 99.00 & 88.29 & 11.77 & 11.77 \\ 
\hdashline
MAE ViT-L/16 & 1 & 68.60 & 98.10 & 91.70 & 94.20 & 85.70 & 91.90 & 64.50 & 97.20 & 96.00 & 71.30 & 99.70 & 75.10 & 72.30 & 84.70 & 98.45 & 89.65 & \textbf{10.34} & \textbf{10.35} \\
DINOv2 ViT-L/14-Reg & 1 & 75.60 & 97.20 & 92.80 & 94.80 & 81.60 & 93.40 & 30.80 & 99.60 & 99.75 & 62.70 & 99.74 & 76.90 & 74.90 & 86.10 & 99.85 & 88.78 & \underline{11.32} & \underline{11.26} \\ 
\hdashline
MAE ViT-L/16 & 15 & 80.50 & 99.20 & 94.10 & 93.50 & 90.30 & 95.80 & 77.60 & 99.70 & 99.70 & 77.30 & 99.92 & 85.10 & 81.50 & 84.20 & 99.25 & 93.16 & 6.88 & 6.81 \\
DINOv2 ViT-L/14-Reg & 11 & 85.10 & 98.60 & 94.30 & 95.40 & 89.70 & 97.20 & 67.50 & 99.60 & 99.85 & 92.40 & 99.86 & 86.80 & 88.10 & 89.00 & 99.65 & 94.38 & \textbf{5.63} & \textbf{5.64} \\
\hline
\end{tabular}
}
\end{table*}

We selected the DINOv2 - ViT-L/14-Reg (chosen for its balance between performance and model size) to assess potential enhancements. Simple linear adaptors were utilized to reduce feature dimensionality and enable feature concatenation. Additionally, dropout was applied to mitigate overfitting. The results are presented in Tab.~\ref{tab:approach_1_abl}. The optimal configuration involves using dropout alongside linear adaptors and feature concatenation.

We applied the optimal configuration on the selected backbones and then compared their performances. The results are displayed in Tab.~\ref{tab:comparison}. DINOv2 and MAE clearly outperformed EfficientNetV2 and DeiT III, and it surpassed EVA-CLIP despite the latter's pre-training on a larger dataset with rich annotations (image-text pairs). Between the two SSL models, MAE was slightly better than DINOv2. These results underscore the advantage of using SSL for pre-training, enabling the learning of superior representations applicable to multiple tasks.

\subsection{Approach 2: Fine-tuning final transformer blocks}
\label{sec:eval_approach_2}
We fine-tuned the last blocks (and the tokens in the case of transformers) of the selected backbones and compared their performances. Details are shown in Tab.~\ref{tab:comparison}. Compared to \textbf{Approach 1}, all backbones gained better results, with EVA-CLIP being the closest competitor with the SSL backbones. Nevertheless, DINOv2 and MAE remained the top performers. To narrow its gaps with DINOv2 and MAE, EVA-CLIP would need to be pre-trained with a vast dataset featuring rich annotations---a costly endeavor compared to DINOv2 and MAE, which was pre-trained on a substantially smaller dataset without any annotations. Given the same architecture (DeiT III versus DINOv2 and MAE), the performance gaps are significant, thanks to the training receipts. Overall, these results again underscore the significant advantage of using SSL for pre-training ViTs.

\begin{figure}[t]
\centering
\includegraphics[width=0.90\columnwidth]{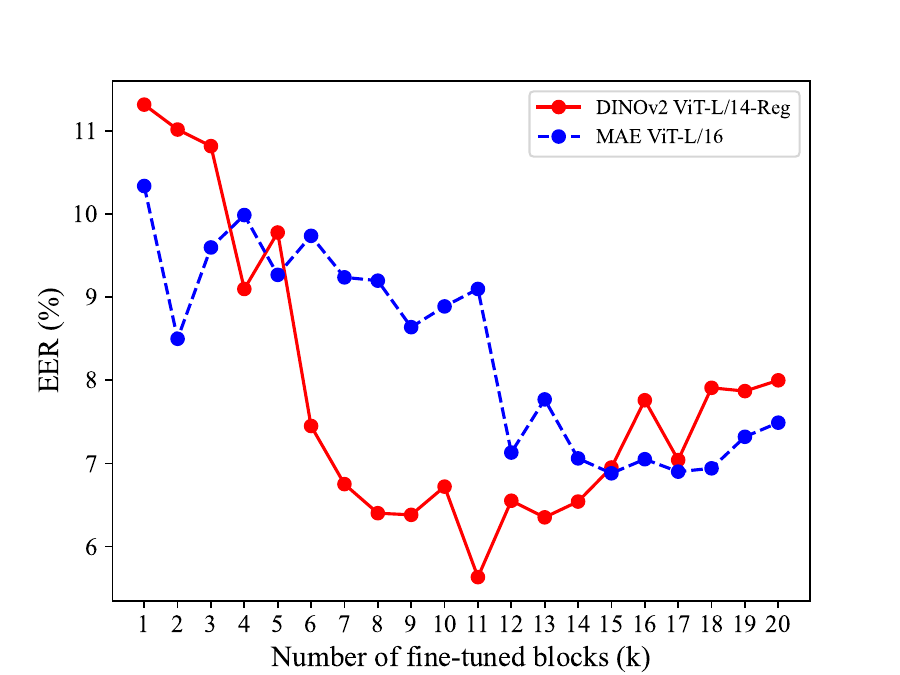}
\caption{Ablation study on the relationship between the number of fine-tuned transformer blocks ($k$) and the EER in \textbf{Approach 2}.}
\label{fig:blocks_eers}
\end{figure}

Next, we conducted an ablation study to determine the optimal number of $k$ final blocks required for fine-tuning. It is important to note that different ViT backbones have varying numbers of blocks; for example, DINOv2 - ViT-L/14-Reg has 24 blocks. The results are visualized in Fig.~\ref{fig:blocks_eers}. If $k$ is small, the model may not adapt adequately to the new task, resulting in underfitting. Conversely, a large $k$ can lead to overfitting, especially with a small fine-tuning dataset. For DINOv2, the optimal $k$ is about half of the total blocks (11), while for MAE, it is three-fifths (15). The EERs decreased from 11.32\% to 5.63\% (-5.69\%) for DINOv2 and from 10.34\% to 6.88\% (-3.46\%) for MAE. When fine-tuning only the last block, MAE had about 1\% better performance compared to DINOv2. However, with the optimal $k$ blocks, DINOv2 surpassed MAE, improving by about 2\% compared to MAE's 1\%.

Regarding the breakdown of results for \textbf{both approaches}, the real parts of FF++, Google DFD, DF-TIMIT, YouTube-DF, and DFDC are the most challenging subsets for detection. This difficulty may be due to the low quality of the deepfake media and diversity of deepfake methods in these subsets. Low quality can destroy artifacts, making detection harder. Fine-tuning with optimal $k$ blocks improved performance for \textbf{Approach 2}, but the real part of Google DFD still remained the most challenging.

\subsection{Cross-dataset detection}
\label{sec:cross_dataset}
\begin{table*}[t]
\centering
\caption{Comparison of performance between various ConvNet and transformer architectures on the unseen test set, comprising images generated or manipulated by diffusion-based methods. $k$ denotes the number of final blocks utilized for feature extraction in \textbf{Approach 1} while signifies the number of fine-tuned blocks in \textbf{Approach 2}.}
\label{tab:unseen_cmp}
\resizebox{\textwidth}{!}{
\begin{tabular}{l|c:c:cccccccccc} 
\hline
\multicolumn{1}{c|}{\vcell{\textbf{Model}}} & \vcell{$\mathbf{k}$} & \vcell{\textbf{Threshold}} & \vcell{\textbf{Real}} & \vcell{\begin{tabular}[b]{@{}c@{}}\textbf{Repaint}\\\textbf{P2}\end{tabular}} & \vcell{\begin{tabular}[b]{@{}c@{}}\textbf{Repaint}\\\textbf{LDM}\end{tabular}} & \vcell{\textbf{LaMa}} & \vcell{\textbf{Pluralistic}} & \vcell{\textbf{Acc.}} & \vcell{\textbf{TPR}} & \vcell{\textbf{TNR}} & \vcell{\textbf{EER}} & \vcell{\textbf{HTER}} \\[-\rowheight]
\multicolumn{1}{c|}{\printcelltop} & \printcelltop & \printcelltop & \printcelltop & \printcelltop & \printcelltop & \printcelltop & \printcelltop & \printcelltop & \printcelltop & \printcelltop & \printcelltop & \printcelltop \\ 
\hline
\rowcolor[rgb]{0.753,0.753,0.753} \textbf{Approach 1} & & & & & & & & & & & & \\
EfficientNetV2 Large & 4 & 0.6355 & 47.89 & 52.89 & 55.78 & 49.11 & 56.67 & 52.47 & 47.89 & 53.61 & 49.22 & 49.25 \\
DeiT III L/16-LayerScale & 4 & 0.9983 & 52.89 & 52.67 & 56.44 & 44.44 & 52.67 & 51.82 & 52.89 & 51.56 & 47.83 & 47.78 \\
EVA-02-CLIP-L/14~ & 4 & 0.0737 & 63.44 & 43.00 & 52.44 & 31.22 & 51.22 & 48.27 & 63.44 & 44.47 & 45.75 & 46.04 \\
\hdashline
MAE ViT-L/16 & 4 & 0.9385 & 56.00 & 53.22 & 62.78 & 19.33 & 64.67 & 51.20 & 56.00 & 50.00 & 47.31 & 47.00 \\
DINOv2 ViT-L/14-Reg & 4 & 0.0759 & 60.44 & 53.00 & 66.89 & 43.11 & 70.89 & 58.87 & 60.44 & 58.47 & \textbf{40.67} & \textbf{40.54} \\ 
\hline
\rowcolor[rgb]{0.753,0.753,0.753} \textbf{Approach 2} & & & & & & & & & & & & \\
EfficientNetV2 Large & 1 & 0.5479 & 63.00 & 50.22 & 53.89 & 65.78 & 64.89 & 59.56 & 63.00 & 58.69 & 39.58 & 39.15 \\
DeiT III L/16-LayerScale & 1 & 0.9999 & 56.00 & 58.56 & 69.56 & 39.56 & 69.56 & 58.64 & 56.00 & 59.31 & 42.56 & 42.35 \\
EVA-02-CLIP-L/14~ & 1 & 0.9999 & 45.44 & 71.11 & 83.44 & 12.22 & 82.11 & 58.87 & 45.44 & 62.22 & 45.20 & 46.17 \\
\hdashline
MAE ViT-L/16 & 1 & 0.1769 & 65.44 & 47.11 & 58.22 & 13.67 & 71.00 & 51.09 & 65.44 & 47.50 & 44.22 & 43.53 \\
DINOv2 ViT-L/14-Reg & 1 & 0.9980 & 50.78 & 70.22 & 78.22 & 65.00 & 86.78 & 70.20 & 50.78 & 75.06 & \textbf{36.28} & \textbf{37.08} \\
\hdashline
MAE ViT-L/16 & 15 & 0.8948 & 69.89 & 50.78 & 68.89 & 22.56 & 76.44 & 57.71 & 69.89 & 54.67 & 37.56 & 37.72 \\
DINOv2 ViT-L/14-Reg & 11 & 0.7418 & 70.22 & 53.22 & 73.22 & 93.00 & 74.56 & 72.84 & 70.22 & 73.50 & \textbf{27.61} & \textbf{28.14} \\
\hline
\end{tabular}}
\end{table*}

In this experiment, we assessed the generalizability of the detectors in detecting unseen deepfakes. The scenario presented a robust challenge, as there were no diffusion images in the training set. The classification thresholds were recalibrated using the unseen validation set. The results are presented in Tab.~\ref{tab:unseen_cmp}. Notably, there were drops in the performance of all models, with the best one going from 11.32\% to 27.61\% in terms of EER. Overall, \textbf{Approach 2} consistently outperformed \textbf{Approach 1}. Within \textbf{Approach 2}, EfficientNetV2 exhibited better generalizability compared to other supervised pre-trained transformers.

DINOv2 is the absolute winner and clearly outperformed MAE. This can be explained by the SSL pre-training phase; DINOv2 employs strong data augmentations, while MAE uses little or none, making MAE less robust against unseen distributions. These results indicate that having the right SSL training strategy greatly enhances deepfake detection performance and is crucial for improving the generalizability of the backbone.

\subsection{Visualization and explainability}
\label{sec:visualization}
\begin{figure*}[t!]
\centering
\includegraphics[width=144mm]{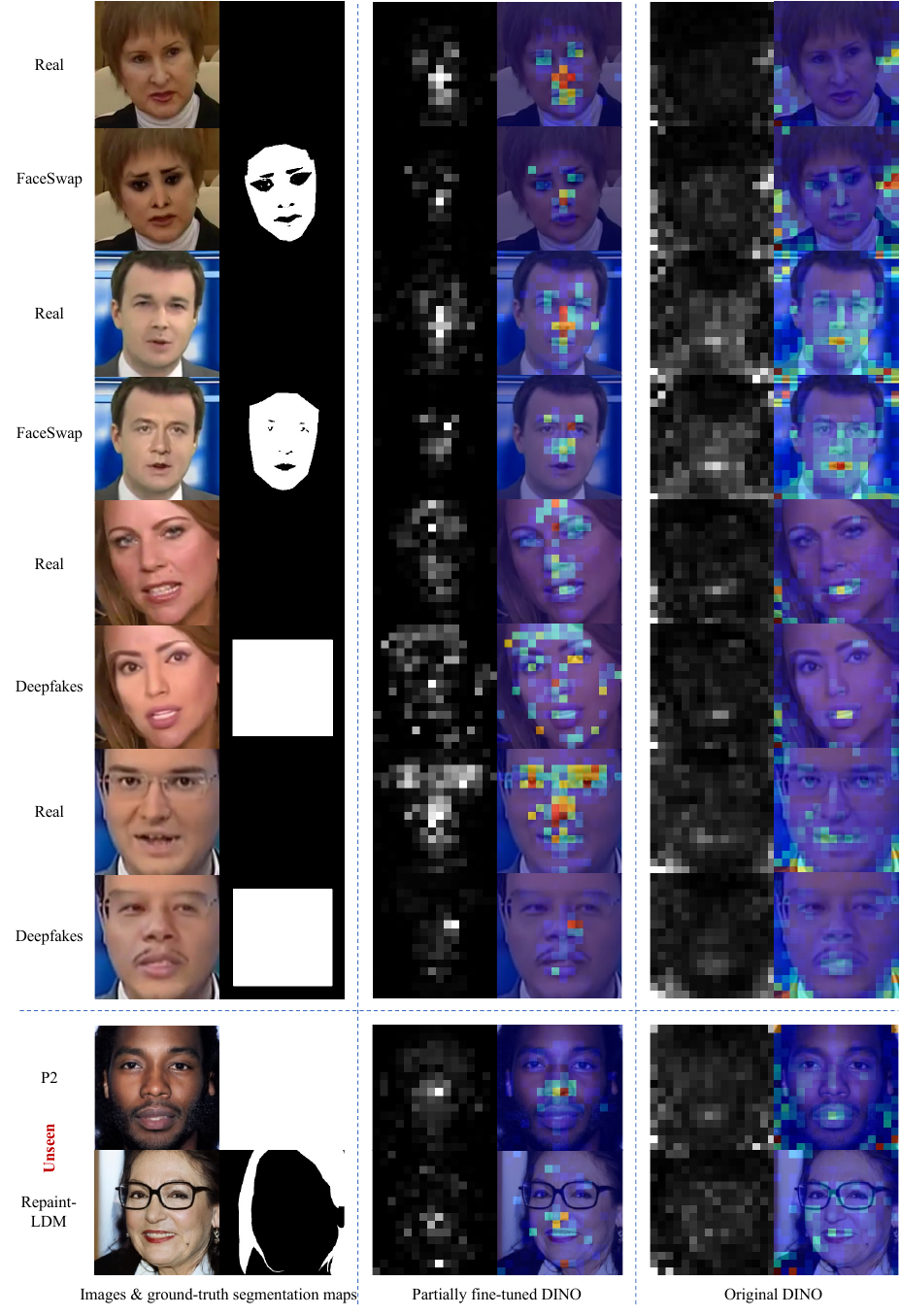}
\caption{Visualization of averages of final block's multi-head attention maps of the partially fine-tuned DINOv2 - ViT-L/14-Reg from \textbf{Approach 2}, compared with those from its original pre-trained version. The training dataset includes Deepfakes and FaceSwap, while P2 and Repaint-LDM are unseen methods. Best viewed in color.}
\label{fig:visualization}
\end{figure*}

With \textbf{Approach 2}, we can naturally visualize the focus areas of the ViT-based models using attention weights. To simplify the process, we computed the average of the attention maps from all attention heads directed toward the CLS token. We chose DINOv2 - ViT-L/14-Reg and randomly selected images per category for visualization to avoid cherry-picking. The results are depicted in Fig.~\ref{fig:visualization}. To highlight the efficacy of fine-tuning, we compared the outcomes with those of the corresponding frozen original model. The partially fine-tuned model primarily directed its attention to the forehead, eyes, nose, and mouth to assess the authenticity of the input image. This behavior closely mirrors human intuition in deepfake detection, as deepfake artifacts frequently manifest in these regions. Notably, the original version of DINO did not possess this ability. Even when presented with unseen deepfakes, the fine-tuned model consistently prioritized these areas. This explains the model's failure to detect deepfakes generated by Repaint-LDM, where the modification occurs in the hair region. In summary, such visualizations play a crucial role in deepfake detection, enhancing the interpretability of the results. The partially fine-tuned DINO model excelled in this regard.

\section{Conclusion and future work}
In this study, we explored two strategies for utilizing SSL pre-trained ViTs--specifically DINOs and MAE--as feature extractors for deepfake detection. The first approach involved utilizing frozen ViT backbones to extract multi-level features, while the second approach entailed partial fine-tuning on the final $k$ blocks. Through extensive experimentation, we found that with a suitable SSL pre-training strategy, the fine-tuning approach demonstrated superior performance and interpretability, particularly through attention mechanisms to visualize the focused areas. Our findings provide valuable insights for the digital forensic community regarding the utilization of SSL pre-trained ViTs as feature extractors, a relatively underexplored area in the literature of deepfake detection.

Future work will primarily concentrate on forensic localization using DINOs without utilizing segmentation ground-truths during training. Additionally, efforts will be directed toward enhancing the generalizability of the models and exploring the potential of SSL on unlabeled deepfake datasets.

\section*{Acknowledgements}
This work was partially supported by JSPS KAKENHI Grants JP21H04907 and JP24H00732, by JST CREST Grants JPMJCR18A6 and JPMJCR20D3 including AIP challenge program, by JST AIP Acceleration Grant JPMJCR24U3 Japan, and by the project for the development and demonstration of countermeasures against disinformation and misinformation on the Internet with the Ministry of Internal Affairs and Communications of Japan.

{\small
\bibliographystyle{ieee}
\bibliography{references}

\begin{thebibliography}{10}\itemsep=-1pt

\bibitem{achiam2023gpt}
J.~Achiam, S.~Adler, S.~Agarwal, L.~Ahmad, I.~Akkaya, F.~L. Aleman, D.~Almeida,
  J.~Altenschmidt, S.~Altman, S.~Anadkat, et~al.
\newblock {GPT-4} technical report.
\newblock {\em arXiv preprint arXiv:2303.08774}, 2023.

\bibitem{kakaobrain2022coyo-700m}
M.~Byeon, B.~Park, H.~Kim, S.~Lee, W.~Baek, and S.~Kim.
\newblock {COYO-700M}: Image-text pair dataset.
\newblock \url{https://github.com/kakaobrain/coyo-dataset}, 2022.

\bibitem{cardenuto2023age}
J.~P. Cardenuto, J.~Yang, R.~Padilha, R.~Wan, D.~Moreira, H.~Li, S.~Wang,
  F.~Andal{\'o}, S.~Marcel, A.~Rocha, et~al.
\newblock The age of synthetic realities: Challenges and opportunities.
\newblock {\em APSIPA Transactions on Signal and Information Processing},
  12(1), 2023.

\bibitem{caron2020unsupervised}
M.~Caron, I.~Misra, J.~Mairal, P.~Goyal, P.~Bojanowski, and A.~Joulin.
\newblock Unsupervised learning of visual features by contrasting cluster
  assignments.
\newblock {\em Advances in NeurIPS}, 33:9912--9924, 2020.

\bibitem{caron2021emerging}
M.~Caron, H.~Touvron, I.~Misra, H.~J{\'e}gou, J.~Mairal, P.~Bojanowski, and
  A.~Joulin.
\newblock Emerging properties in self-supervised vision transformers.
\newblock In {\em ICCV}, pages 9650--9660, 2021.

\bibitem{choi2022perception}
J.~Choi, J.~Lee, C.~Shin, S.~Kim, H.~Kim, and S.~Yoon.
\newblock Perception prioritized training of diffusion models.
\newblock In {\em CVPR}, pages 11472--11481, 2022.

\bibitem{choi2018stargan}
Y.~Choi, M.~Choi, M.~Kim, J.-W. Ha, S.~Kim, and J.~Choo.
\newblock {StarGAN}: Unified generative adversarial networks for multi-domain
  image-to-image translation.
\newblock In {\em CVPR}, pages 8789--8797, 2018.

\bibitem{choi2020stargan}
Y.~Choi, Y.~Uh, J.~Yoo, and J.-W. Ha.
\newblock {StarGAN} v2: Diverse image synthesis for multiple domains.
\newblock In {\em CVPR}, pages 8188--8197, 2020.

\bibitem{chollet2017xception}
F.~Chollet.
\newblock Xception: Deep learning with depthwise separable convolutions.
\newblock In {\em CVPR}, pages 1251--1258, 2017.

\bibitem{chung2018voxceleb2}
J.~S. Chung, A.~Nagrani, and A.~Zisserman.
\newblock {VoxCeleb2}: Deep speaker recognition.
\newblock In {\em {INTERSPEECH}}, pages 1086--1090, 2018.

\bibitem{cocchi2023unveiling}
F.~Cocchi, L.~Baraldi, S.~Poppi, M.~Cornia, L.~Baraldi, and R.~Cucchiara.
\newblock Unveiling the impact of image transformations on deepfake detection:
  An experimental analysis.
\newblock In {\em ICIAP}, pages 345--356. Springer, 2023.

\bibitem{coccomini2022combining}
D.~A. Coccomini, N.~Messina, C.~Gennaro, and F.~Falchi.
\newblock Combining efficientnet and vision transformers for video deepfake
  detection.
\newblock In {\em ICIAP}, pages 219--229. Springer, 2022.

\bibitem{darcet2024vision}
T.~Darcet, M.~Oquab, J.~Mairal, and P.~Bojanowski.
\newblock Vision transformers need registers.
\newblock In {\em ICLR}, 2024.

\bibitem{das2024limited}
S.~Das, T.~Jain, D.~Reilly, P.~Balaji, S.~Karmakar, S.~Marjit, X.~Li, A.~Das,
  and M.~S. Ryoo.
\newblock Limited data, unlimited potential: A study on vits augmented by
  masked autoencoders.
\newblock In {\em WACV}, pages 6878--6888, 2024.

\bibitem{deng2009imagenet}
J.~Deng, W.~Dong, R.~Socher, L.-J. Li, K.~Li, and L.~Fei-Fei.
\newblock {ImageNet}: A large-scale hierarchical image database.
\newblock In {\em CVPR}, pages 248--255. IEEE, 2009.

\bibitem{dolhansky2020deepfake}
B.~Dolhansky, J.~Bitton, B.~Pflaum, J.~Lu, R.~Howes, M.~Wang, and C.~C. Ferrer.
\newblock The deepfake detection challenge ({DFDC}) dataset.
\newblock {\em arXiv preprint arXiv:2006.07397}, 2020.

\bibitem{dosovitskiy2020image}
A.~Dosovitskiy, L.~Beyer, A.~Kolesnikov, D.~Weissenborn, X.~Zhai,
  T.~Unterthiner, M.~Dehghani, M.~Minderer, G.~Heigold, S.~Gelly, et~al.
\newblock An image is worth 16x16 words: Transformers for image recognition at
  scale.
\newblock In {\em ICLR}, 2020.

\bibitem{googledfd}
N.~Dufour and A.~Gully.
\newblock Contributing data to deepfake detection research.
\newblock
  \url{https://ai.googleblog.com/2019/09/contributing-data-to-deepfake-detection.html},
  9 2019.

\bibitem{guan2022delving}
J.~Guan, H.~Zhou, Z.~Hong, E.~Ding, J.~Wang, C.~Quan, and Y.~Zhao.
\newblock Delving into sequential patches for deepfake detection.
\newblock {\em Advances in NeurIPS}, 35:4517--4530, 2022.

\bibitem{he2022masked}
K.~He, X.~Chen, S.~Xie, Y.~Li, P.~Doll{\'a}r, and R.~Girshick.
\newblock Masked autoencoders are scalable vision learners.
\newblock In {\em CVPR}, pages 16000--16009, 2022.

\bibitem{he2020momentum}
K.~He, H.~Fan, Y.~Wu, S.~Xie, and R.~Girshick.
\newblock Momentum contrast for unsupervised visual representation learning.
\newblock In {\em CVPR}, pages 9729--9738, 2020.

\bibitem{heo2023deepfake}
Y.-J. Heo, W.-H. Yeo, and B.-G. Kim.
\newblock Deepfake detection algorithm based on improved vision transformer.
\newblock {\em Applied Intelligence}, 53(7):7512--7527, 2023.

\bibitem{hinton2015distilling}
G.~Hinton, O.~Vinyals, and J.~Dean.
\newblock Distilling the knowledge in a neural network.
\newblock {\em arXiv preprint arXiv:1503.02531}, 2015.

\bibitem{karras2018progressive}
T.~Karras, T.~Aila, S.~Laine, and J.~Lehtinen.
\newblock Progressive growing of gans for improved quality, stability, and
  variation.
\newblock In {\em ICLR}, 2018.

\bibitem{karras2019style}
T.~Karras, S.~Laine, and T.~Aila.
\newblock A style-based generator architecture for generative adversarial
  networks.
\newblock In {\em CVPR}, pages 4401--4410, 2019.

\bibitem{karras2020analyzing}
T.~Karras, S.~Laine, M.~Aittala, J.~Hellsten, J.~Lehtinen, and T.~Aila.
\newblock Analyzing and improving the image quality of {StyleGAN}.
\newblock In {\em CVPR}, pages 8110--8119, 2020.

\bibitem{kenton2019bert}
J.~D. M.-W.~C. Kenton and L.~K. Toutanova.
\newblock {BERT}: Pre-training of deep bidirectional transformers for language
  understanding.
\newblock In {\em NAACL-HLT}, pages 4171--4186, 2019.

\bibitem{khan2021video}
S.~A. Khan and H.~Dai.
\newblock Video transformer for deepfake detection with incremental learning.
\newblock In {\em ACM MM}, pages 1821--1828, 2021.

\bibitem{korshunov2018deepfakes}
P.~Korshunov and S.~Marcel.
\newblock Deepfakes: a new threat to face recognition? assessment and
  detection.
\newblock {\em arXiv preprint arXiv:1812.08685}, 2018.

\bibitem{kukanov2020cost}
I.~Kukanov, J.~Karttunen, H.~Sillanp{\"a}{\"a}, and V.~Hautam{\"a}ki.
\newblock Cost sensitive optimization of deepfake detector.
\newblock In {\em APSIPA ASC}, pages 1300--1303. IEEE, 2020.

\bibitem{li2020celeb}
Y.~Li, X.~Yang, P.~Sun, H.~Qi, and S.~Lyu.
\newblock Celeb-df: A large-scale challenging dataset for deepfake forensics.
\newblock In {\em CVPR}, pages 3207--3216, 2020.

\bibitem{lin2023deepfake}
H.~Lin, W.~Huang, W.~Luo, and W.~Lu.
\newblock Deepfake detection with multi-scale convolution and vision
  transformer.
\newblock {\em Digital Signal Processing}, 134:103895, 2023.

\bibitem{liu2024forgery}
H.~Liu, Z.~Tan, C.~Tan, Y.~Wei, Y.~Zhao, and J.~Wang.
\newblock Forgery-aware adaptive transformer for generalizable synthetic image
  detection.
\newblock In {\em CVPR}, 2023.

\bibitem{nguyen2023close}
H.~H. Nguyen, J.~Yamagishi, and I.~Echizen.
\newblock How close are other computer vision tasks to deepfake detection?
\newblock In {\em IJCB}, pages 1--10. IEEE, 2023.

\bibitem{ojha2023towards}
U.~Ojha, Y.~Li, and Y.~J. Lee.
\newblock Towards universal fake image detectors that generalize across
  generative models.
\newblock In {\em CVPR}, pages 24480--24489, 2023.

\bibitem{oquab2023dinov2}
M.~Oquab, T.~Darcet, T.~Moutakanni, H.~V. Vo, M.~Szafraniec, V.~Khalidov,
  P.~Fernandez, D.~HAZIZA, F.~Massa, A.~El-Nouby, et~al.
\newblock Dinov2: Learning robust visual features without supervision.
\newblock {\em Transactions on Machine Learning Research}, 2023.

\bibitem{radford2021learning}
A.~Radford, J.~W. Kim, C.~Hallacy, A.~Ramesh, G.~Goh, S.~Agarwal, G.~Sastry,
  A.~Askell, P.~Mishkin, J.~Clark, et~al.
\newblock Learning transferable visual models from natural language
  supervision.
\newblock In {\em ICML}, pages 8748--8763. PMLR, 2021.

\bibitem{radford2019language}
A.~Radford, J.~Wu, R.~Child, D.~Luan, D.~Amodei, I.~Sutskever, et~al.
\newblock Language models are unsupervised multitask learners.
\newblock {\em OpenAI blog}, 1(8):9, 2019.

\bibitem{rana2022deepfake}
M.~S. Rana, M.~N. Nobi, B.~Murali, and A.~H. Sung.
\newblock Deepfake detection: A systematic literature review.
\newblock {\em IEEE access}, 10:25494--25513, 2022.

\bibitem{rombach2022high}
R.~Rombach, A.~Blattmann, D.~Lorenz, P.~Esser, and B.~Ommer.
\newblock High-resolution image synthesis with latent diffusion models.
\newblock In {\em CVPR}, pages 10684--10695, 2022.

\bibitem{rossler2019faceforensics++}
A.~Rossler, D.~Cozzolino, L.~Verdoliva, C.~Riess, J.~Thies, and M.~Nie{\ss}ner.
\newblock Faceforensics++: Learning to detect manipulated facial images.
\newblock In {\em ICCV}, pages 1--11, 2019.

\bibitem{sanderson2009multi}
C.~Sanderson and B.~C. Lovell.
\newblock Multi-region probabilistic histograms for robust and scalable
  identity inference.
\newblock In {\em ICB}, pages 199--208. Springer, 2009.

\bibitem{schuhmann2022laion}
C.~Schuhmann, R.~Beaumont, R.~Vencu, C.~Gordon, R.~Wightman, M.~Cherti,
  T.~Coombes, A.~Katta, C.~Mullis, M.~Wortsman, et~al.
\newblock {LAION-5B}: An open large-scale dataset for training next generation
  image-text models.
\newblock {\em Advances in NeurIPS}, 35:25278--25294, 2022.

\bibitem{sun2023eva}
Q.~Sun, Y.~Fang, L.~Wu, X.~Wang, and Y.~Cao.
\newblock Eva-clip: Improved training techniques for clip at scale.
\newblock {\em arXiv preprint arXiv:2303.15389}, 2023.

\bibitem{suvorov2022resolution}
R.~Suvorov, E.~Logacheva, A.~Mashikhin, A.~Remizova, A.~Ashukha, A.~Silvestrov,
  N.~Kong, H.~Goka, K.~Park, and V.~Lempitsky.
\newblock Resolution-robust large mask inpainting with fourier convolutions.
\newblock In {\em WACV}, pages 2149--2159, 2022.

\bibitem{tan2019efficientnet}
M.~Tan and Q.~Le.
\newblock Efficientnet: Rethinking model scaling for convolutional neural
  networks.
\newblock In {\em ICML}, pages 6105--6114. PMLR, 2019.

\bibitem{tan2021efficientnetv2}
M.~Tan and Q.~Le.
\newblock Efficientnetv2: Smaller models and faster training.
\newblock In {\em ICML}, pages 10096--10106. PMLR, 2021.

\bibitem{tolosana2020deepfakes}
R.~Tolosana, R.~Vera-Rodriguez, J.~Fierrez, A.~Morales, and J.~Ortega-Garcia.
\newblock Deepfakes and beyond: A survey of face manipulation and fake
  detection.
\newblock {\em Information Fusion}, 64:131--148, 2020.

\bibitem{touvron2021training}
H.~Touvron, M.~Cord, M.~Douze, F.~Massa, A.~Sablayrolles, and H.~J{\'e}gou.
\newblock Training data-efficient image transformers \& distillation through
  attention.
\newblock In {\em ICML}, pages 10347--10357. PMLR, 2021.

\bibitem{touvron2022deit}
H.~Touvron, M.~Cord, and H.~J{\'e}gou.
\newblock {DeiT III}: Revenge of the {VIT}.
\newblock In {\em European conference on computer vision}, pages 516--533.
  Springer, 2022.

\bibitem{vaswani2017attention}
A.~Vaswani, N.~Shazeer, N.~Parmar, J.~Uszkoreit, L.~Jones, A.~N. Gomez,
  {\L}.~Kaiser, and I.~Polosukhin.
\newblock Attention is all you need.
\newblock {\em Advances in NIPS}, 30, 2017.

\bibitem{wang2022m2tr}
J.~Wang, Z.~Wu, W.~Ouyang, X.~Han, J.~Chen, Y.-G. Jiang, and S.-N. Li.
\newblock M2tr: Multi-modal multi-scale transformers for deepfake detection.
\newblock In {\em ICMR}, pages 615--623, 2022.

\bibitem{wang2023deep}
T.~Wang, H.~Cheng, K.~P. Chow, and L.~Nie.
\newblock Deep convolutional pooling transformer for deepfake detection.
\newblock {\em ACM Transactions on Multimedia Computing, Communications and
  Applications}, 19(6):1--20, 2023.

\bibitem{wu2019relgan}
P.-W. Wu, Y.-J. Lin, C.-H. Chang, E.~Y. Chang, and S.-W. Liao.
\newblock {RelGAN}: Multi-domain image-to-image translation via relative
  attributes.
\newblock In {\em ICCV}, pages 5914--5922, 2019.

\bibitem{zhao2023istvt}
C.~Zhao, C.~Wang, G.~Hu, H.~Chen, C.~Liu, and J.~Tang.
\newblock Istvt: interpretable spatial-temporal video transformer for deepfake
  detection.
\newblock {\em IEEE Transactions on Information Forensics and Security},
  18:1335--1348, 2023.

\bibitem{zheng2019pluralistic}
C.~Zheng, T.-J. Cham, and J.~Cai.
\newblock Pluralistic image completion.
\newblock In {\em CVPR}, pages 1438--1447, 2019.

\bibitem{tantaru2024weakly}
D.-C. Ț{\^a}nțaru, E.~Oneaț{\u{a}}, and D.~Oneaț{\u{a}}.
\newblock Weakly-supervised deepfake localization in diffusion-generated
  images.
\newblock In {\em WACV}, pages 6258--6268, 2024.

\end{thebibliography}
}

\end{document}